\documentclass[11pt,a4paper]{article}
\usepackage[hyperref]{emnlp-ijcnlp-2019}
\usepackage{times}
\usepackage{latexsym}
\usepackage{graphicx,xcolor,colortbl}
\usepackage{amsmath}
\usepackage{amssymb}
\usepackage{url}
\usepackage{booktabs}
\usepackage{multirow}
\usepackage{verbatim}
\usepackage{textcomp}
\usepackage{tabularx}
\usepackage{makecell}
\usepackage[utf8]{inputenc}
\usepackage{cyrillic}

\newcommand{\cyrrm}{\fontencoding{OT2}\selectfont\textcyrup}

\DeclareMathOperator*{\argmin}{\arg\!\min}

\newcommand\blfootnote[1]{%
  \begingroup
  \renewcommand\thefootnote{}\footnote{#1}%
  \addtocounter{footnote}{-1}%
  \endgroup
}

\makeatletter
\g@addto@macro\normalsize{%
  \setlength\abovedisplayskip{4pt}
  \setlength\belowdisplayskip{4pt}
  \setlength\abovedisplayshortskip{4pt}
  \setlength\belowdisplayshortskip{4pt}
}
\makeatother

\newcommand{\datasetname}{Birds-to-Words}
\newcommand{\modelname}{Neural Naturalist}

\newcommand{\symbolowl}[0]{\includegraphics[height=.02\textwidth]{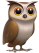}}
\newcommand{\symboldove}[0]{\includegraphics[height=.02\textwidth]{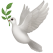}}
\newcommand{\symbolparrot}[0]{\includegraphics[height=.02\textwidth]{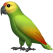}}

\definecolor{website}{rgb}{1.0,0.55,0.0}
\definecolor{jointhighlight}{rgb}{1.0,1.0,0.85}
\definecolor{comphighlight}{rgb}{1.0,0.92,0.91}
\definecolor{decodehighlight}{rgb}{0.64,0.86,0.95}
\definecolor{algohighlight}{rgb}{0.9,0.9,0.9}
\newcommand{\hlcell}{\cellcolor{jointhighlight}}
\newcommand{\hlcompcell}{\cellcolor{comphighlight}}
\newcommand{\hldecodecell}{\cellcolor{decodehighlight}}
\newcommand{\hlalgocell}{\cellcolor{algohighlight}}

\aclfinalcopy 

\newcommand{\authortab}{\hspace{15pt}}
\newcommand{\minitab}{\hspace{12pt}}

\title{Neural Naturalist: Generating Fine-Grained Image Comparisons}

\author{Maxwell Forbes$^{\symbolowl}$ \authortab Christine Kaeser-Chen$^{\symbolparrot}$ \authortab Piyush Sharma$^{\symbolparrot}$ \authortab Serge Belongie$^{\symbolparrot\symboldove}$\vspace{2mm}\\
${\symbolowl}$ University of Washington \authortab
${\symbolparrot}$ Google Research \authortab ${\symboldove}$ Cornell University and Cornell Tech\\
{\tt mbforbes@cs.washington.edu}\\
{\tt \{christinech,piyushsharma\}@google.com}\\
{\tt sjb344@cornell.edu} \vspace{2mm}\\
\textbf{\url{https://mbforbes.github.io/neural-naturalist/}}%
\\
}

\date{}

\begin{document}
\maketitle

\begin{abstract}

We introduce the new \textit{\datasetname} dataset of 41k sentences describing fine-grained differences between photographs of birds. The language collected is highly detailed, while remaining understandable to the everyday observer (e.g., ``heart-shaped face,'' ``squat body''). Paragraph-length descriptions naturally adapt to varying levels of taxonomic and visual distance---drawn from a novel stratified sampling approach---with the appropriate level of detail. We propose a new model called \textit{\modelname} that uses a joint image encoding and comparative module to generate comparative language, and evaluate the results with humans who must use the descriptions to distinguish real images.

Our results indicate promising potential for neural models to explain differences in visual embedding space using natural language, as well as a concrete path for machine learning to aid citizen scientists in their effort to preserve biodiversity.

\end{abstract}
\section{Introduction}

\begin{figure}[t!]

\begin{center}
\includegraphics[width=0.99\linewidth]{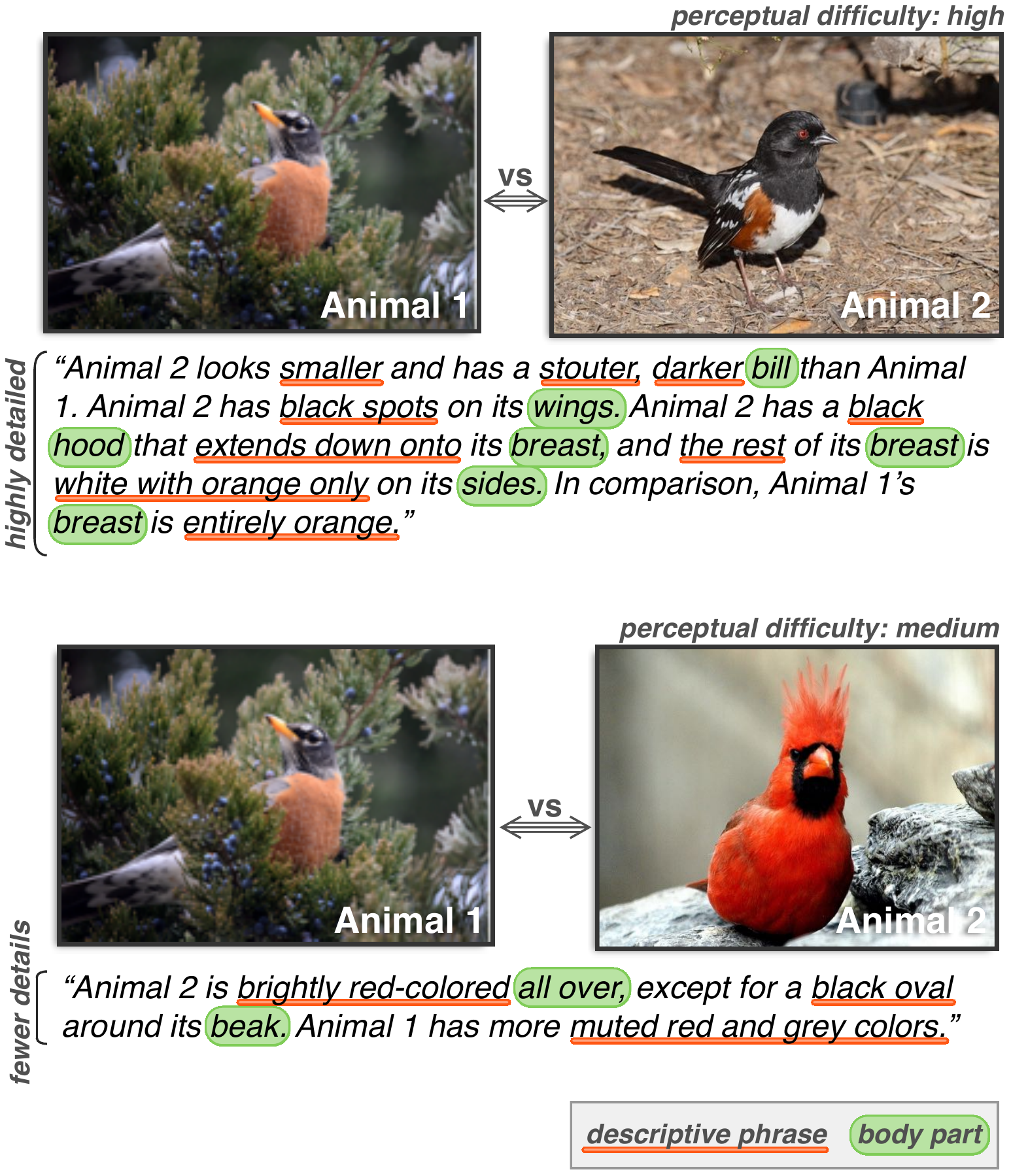}
\end{center}

\caption{
The \datasetname~dataset: comparative descriptions adapt naturally to the appropriate level of detail (orange underlines). A difficult distinction (\textsc{top}) is given a longer and more fined-grained comparison than an easier one (\textsc{bottom}). Annotators organically use everyday language to refer to parts (green highlights).
}
\label{fig:frontpage}

\end{figure}

Humans are adept at making fine-grained comparisons, but sometimes require aid in distinguishing visually similar classes.\blfootnote{${\symbolowl}$ Work done during an internship at Google.}
Take, for example, a citizen science effort like iNaturalist,\footnote{https://www.inaturalist.org} where everyday people photograph wildlife, and the community reaches a consensus on the taxonomic label for each instance.
Many species are visually similar (e.g., Figure~\ref{fig:frontpage}, top), making them difficult for a casual observer to label correctly.
This puts an undue strain on lieutenants of the citizen science community to curate and justify labels for a large number of instances.
While everyone may be capable of making such distinctions visually, non-experts require training to know what to look for.

Field guides exist for the purpose helping people learn how to distinguish between species.
Unfortunately, field guides are costly to create because writing such a guide requires expert knowledge of class-level distinctions.

In this paper, we study the problem of explaining the differences between two images using natural language.
We introduce a new dataset called \textit{\datasetname} of paragraph-length descriptions of the differences between pairs of bird photographs.
We find several benefits from eliciting comparisons: (a) without a guide, annotators naturally break down the subject of the image (e.g., a bird) into pieces understood by the everyday observer (e.g., head, wings, legs); (b) by sampling comparisons from varying visual and taxonomic distances, the language exhibits naturally adaptive granularity of detail based on the distinctions required (e.g., ``red body'' vs ``tiny stripe above its eye''); (c) in contrast to requiring comparisons between categories (e.g., comparing one species vs.~another), non-experts can provide high-quality annotations without needing domain expertise.

We also propose the \textit{\modelname} model architecture for generating comparisons given two images as input. After embedding images into a latent space with a CNN, the model combines the two image representations with a joint encoding and comparative module before passing them to a Transformer decoder. We find that introducing a comparative module---an additional Transformer encoder---over the combined latent image representations yields better generations.

Our results suggest that these classes of neural models can assist in fine-grained visual domains when humans require aid to distinguish closely related instances. Non-experts---such as amateur naturalists trying to tell apart two species---stand to benefit from comparative explanations. Our work approaches this sweet-spot of visual expertise, where any two in-domain images can be compared, and the language is detailed, adaptive to the types of differences observed, and still understandable by laypeople.

Recent work has made impressive progress on context sensitive image captioning. One direction of work uses class labels as context, with the objective of generating captions that distinguish why the image belongs to one class over others \cite{hendricks2016generating,vedantam2017context}. Another choice is to use a second image as context, and generate a caption that distinguishes one image from another. Previous work has studied ways to generalize single-image captions into comparative language \cite{vedantam2017context}, as well as comparing two images with high pixel overlap (e.g., surveillance footage) \cite{jhamtani2018learning}. Our work complements these efforts by studying directly comparative, everyday language on image pairs with no pixel overlap.

\begin{figure}[t]

\begin{center}
\includegraphics[width=0.99\linewidth]{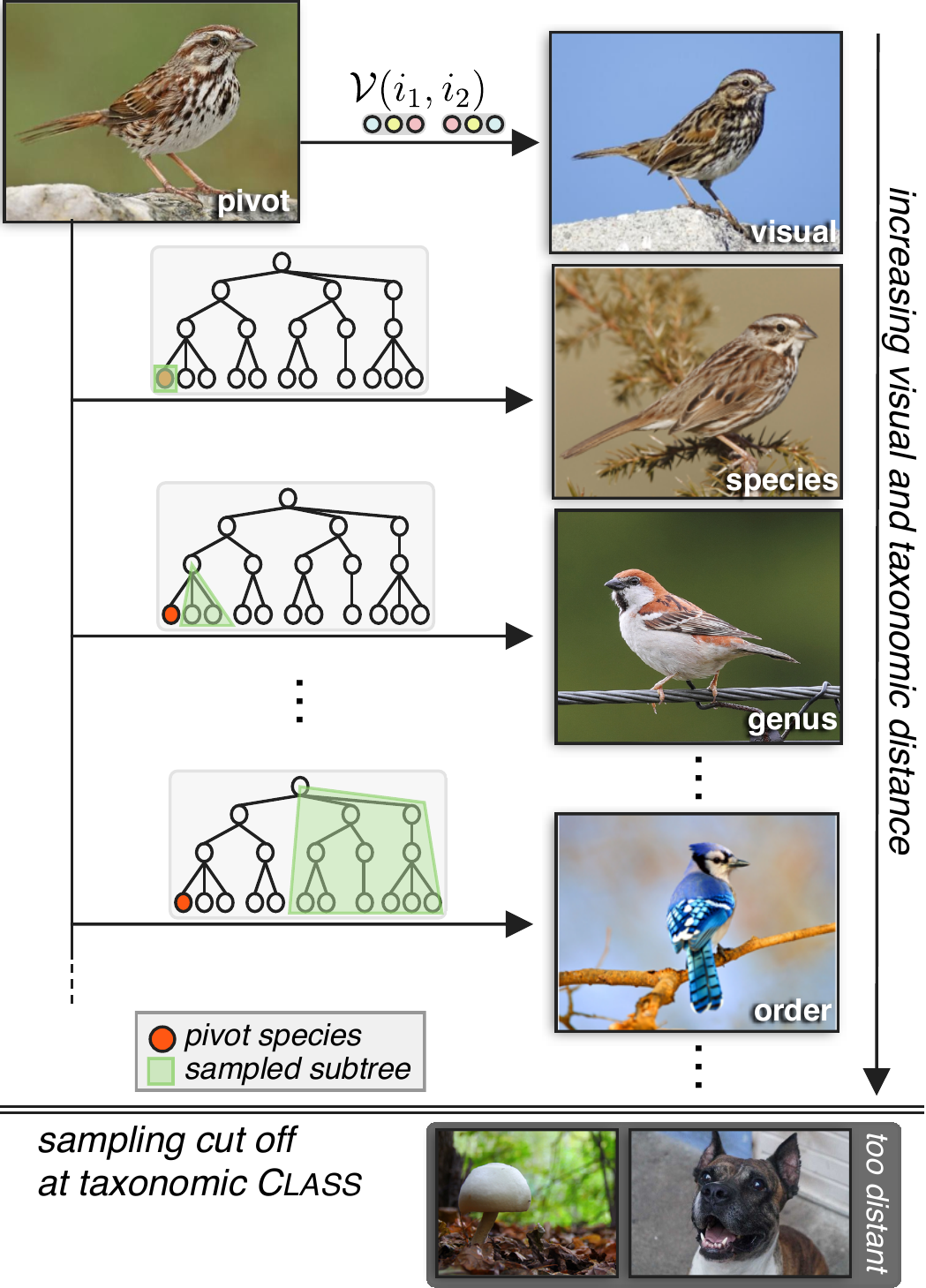}
\end{center}

\caption{
Illustration of pivot-branch stratified sampling algorithm used to construct the \datasetname~dataset. The algorithm harnesses visual and taxonomic distances (increasing vertically) to create a challenging task with board coverage.
}
\label{fig:datasetConstruction}

\end{figure}

Our approach outlines a new way for models to aid humans in making visual distinctions.
The \modelname~model requires two instances as input; these could be, for example, a query image and an image from a candidate class.
By differentiating between these two inputs, a model may help point out subtle distinctions (e.g., one animal has spots on its side), or features that indicate a good match (e.g., only a slight difference in size).
These explanations can aid in understanding both differences between species, as well as variance within instances of a single species.

\begin{table*}[]
\small
\begin{tabularx}{\linewidth}{ l l l l l X }
\toprule
\textbf{} & \textbf{} & \textbf{} & \multicolumn{2}{c}{\textbf{Images}} & \textbf{} \\
\textbf{Dataset} & \textbf{Domain} & \textbf{Lang} & \textbf{Ctx} & \textbf{Cap} & \textbf{Example} \\ \midrule
\makecell{CUB Captions \\ \textit{(R, 2016)}} & Birds & \textsc{m} & 1 & 1 & \textit{``An all black bird with a very long rectrices and relatively dull bill.''} \\ \midrule[0.03em]
\makecell{CUB-Justify\\\textit{(V, 2017)}} & Birds & \textsc{s}  & 7 & 1 & \textit{``The bird has white orbital feathers, a black crown, and yellow tertials.''} \\ \midrule[0.03em]
\makecell{Spot-the-Diff\\\textit{(J\&B, 2018)}} & Surveilance & \textsc{e} & 2 & 1--2 & \textit{''Silver car is gone. Person in a white t shirt appears. 3rd person in the group is gone.''} \\ \midrule[0.03em]
\makecell{\datasetname \\ \textit{(this work)}} & Birds & \textsc{e} & 2 & 2 & \textit{``Animal1 is gray, while animal2 is white. Animal2 has a long, yellow beak, while animal1's beak is shorter and gray. Animal2 appears to be larger than animal1.''} \\ \bottomrule
\end{tabularx}%
\caption{Comparison with recent fine-grained language-and-vision datasets. \textit{Lang} values: \textsc{s} = scientific, \textsc{e} = everyday, \textsc{m} = mixed. \textit{Images Ctx} = number of images shown, \textit{Images Cap} = number of images described in caption. Dataset citations: \textit{R = Reed et al.}, \textit{V = Vedantam et al.}, \textit{J\&B = Jhamtani and Berg-Kirkpatrick}.}
\label{table:datasetComparison}
\end{table*}
\section{\datasetname~Dataset}
\label{sec:dataset}

Our goal is to collect a dataset of tuples $(i_1, i_2, t)$, where $i_1$ and $i_2$ are images, and $t$ is a natural language comparison between the two.
Given a domain $\mathcal{D}$, this collection depends critically on the criteria we use to select image pairs.

If we sample image pairs uniformly at random, we will end up with comparisons encompassing a broad range of phenomena.
For example, two images that are quite different will yield categorical comparisons (\textit{``One is a bird, one is a mushroom.''}).
Alternatively, if the two images are very similar, such as two angles of the same creature, comparisons between them will focus on highly detailed nuances, such as variations in pose.
These phenomena support rich lines of research, such as object classification \cite{deng2009imagenet} and pose estimation \cite{murphy2009head}. 

We aim to land somewhere in the middle.
We wish to consider sets of distinguishable but intimately related pairs.
This sweet spot of visual similarity is akin to the genre of differences studied in fine-grained visual classification \cite{wah2011caltech,krause2013threed}.
We approach this collection with a two-phase data sampling procedure.
We first select \textit{pivot} images by sampling from our full domain uniformly at random.
We then \textit{branch} from these images into a set of secondary images that emphases fine-grained comparisons, but yields broad coverage over the set of sensible relations.
Figure~\ref{fig:datasetConstruction} provides an illustration of our sampling procedure.

\subsection{Domain}

\begin{figure}[]

\begin{center}
\includegraphics[width=0.90\linewidth]{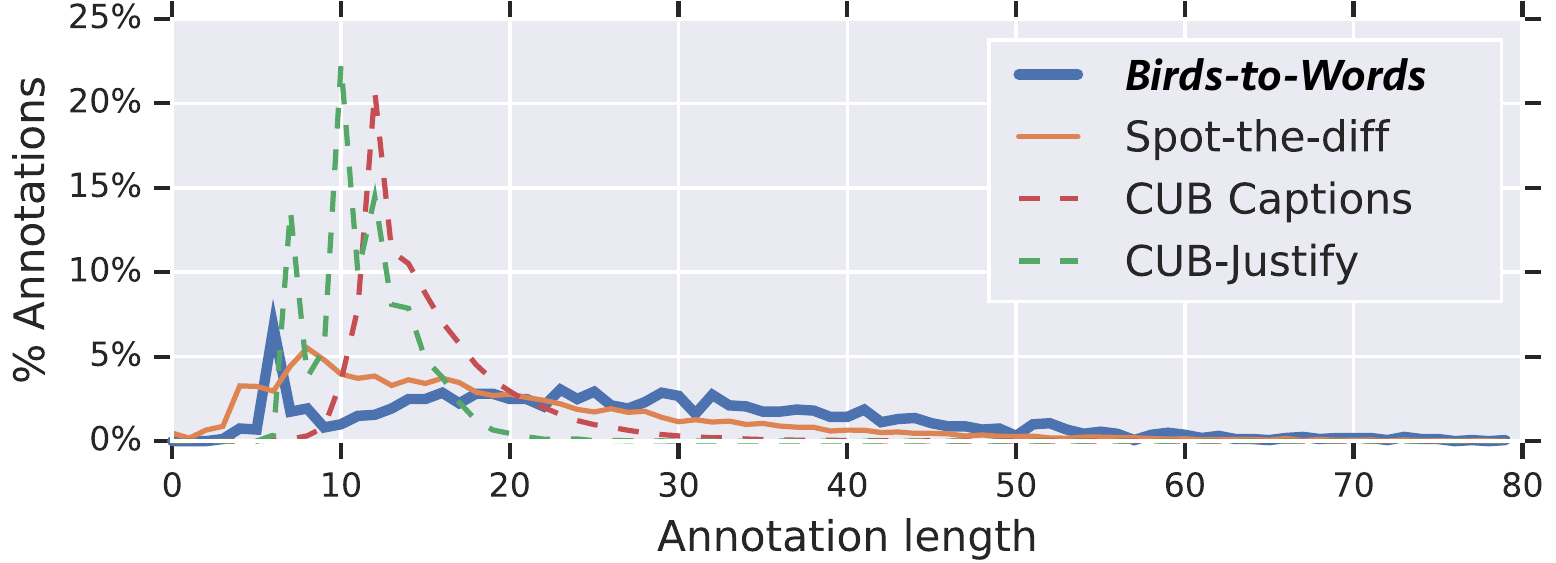}
\end{center}

\centering
\small{
\begin{tabular}{@{}ll@{}}
\toprule
\multicolumn{2}{c}{\datasetname~Dataset}       \\ \midrule
Image pairs           & 3,347              \\
Paragraphs / pair     & 4.8  \\
Paragraphs            & 16,067             \\
Tokens / paragraph    & 32.1 \small{\textsc{mean}} \\ \midrule
Sentences             & 40,969             \\
Sentences / paragraph & 2.6 \small{\textsc{mean}}  \\ \midrule
Clarity rating        & $\geq 4/5$ \\
Train / dev / test    & 80\% / 10\% / 10\% \\ \bottomrule
\end{tabular}
}

\caption{Annotation lengths for compared datasets (\textsc{top}), and statistics for the proposed \datasetname~dataset (\textsc{bottom}).
The \datasetname~dataset has a large mass of long descriptions in comparison to related datasets.
}
\label{figure:dataset}
\end{figure}

We sample images from iNaturalist, a citizen science effort to collect research-grade\footnote{Research-grade observations have met or exceeded iNaturalist's guidelines for community consensus of the taxonomic label for a photograph.} observations of plants and animals in the wild.
We restrict our domain $\mathcal{D}$ to instances labeled under the taxonomic \textsc{class}\footnote{To disambiguate \textit{class}, we use \textsc{class} to denote the taxonomic rank in scientific classification, and simply ``class'' to refer to the machine learning usage of the term as a label in classification.} \textit{Aves} (i.e., birds).
While a broader domain would yield some comparable instances (e.g., \textit{bird} and \textit{dragonfly} share some common body parts), choosing only \textit{Aves} ensures that all instances will be similar enough structurally to be comparable, and avoids the gut reaction comparison pointing out the differences in animal type.
This choice yields 1.7M research-grade images and corresponding taxonomic labels from iNaturalist.
We then perform pivot-branch sampling on this set to choose pairs for annotation.

\begin{figure*}[t!]

\begin{center}
\includegraphics[width=0.99\linewidth]{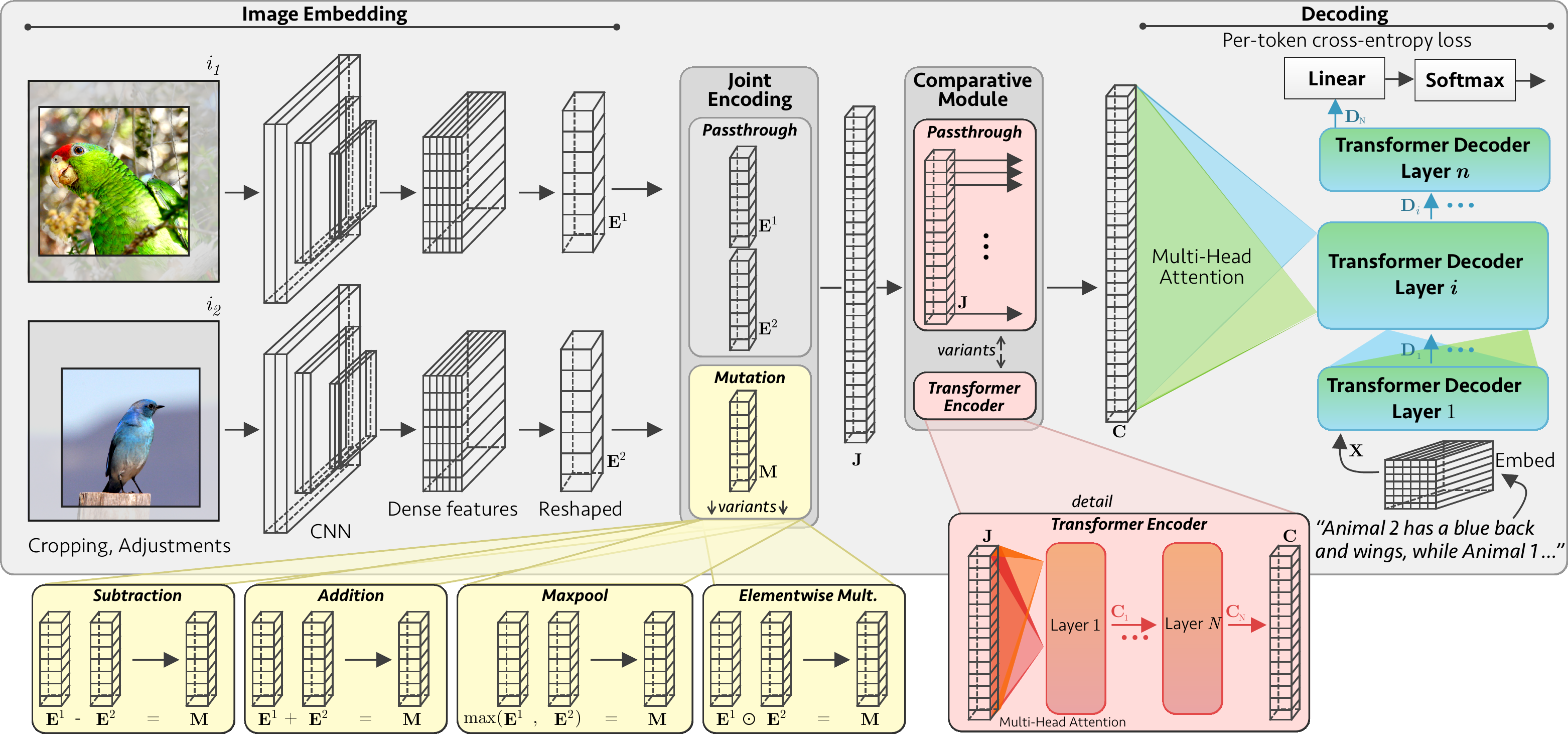}
\end{center}

\caption{The proposed \modelname~model architecture. The multiplicative joint encoding and Transformer-based comparative module yield the best comparisons between images.
}
\label{fig:model}

\end{figure*}

\subsection{Pivot Images}
The \textit{Aves} domain in iNaturalist contains instances of 9k distinct species, with heavy observation bias to more common species (such as the mallard duck).
We uniformly sample species from the set of 9k to help overcome this bias.
In total, we select 405 species and corresponding photographs to use as $i_1$ images.

\subsection{Branching Images}
\label{sec:dataset-branching-images}

We use both a visual similarity measure and taxonomy to sample a set of comparison images $i_2$ branching off from each pivot image $i_1$.
We use a branching factor of $k=12$ from each pivot image.

To capture visually similar images to $i_1$, we employ a similarity function $\mathcal{V}(i_1,i_2)$.
We use an Inception-v4 \cite{szegedy2017inception} network pretrained on ImageNet \cite{deng2009imagenet} and then fine-tuned to perform species classification on all research-grade observations in iNaturalist.
We take the embedding for each image from the last layer of the network before the final softmax.
We perform a k-nearest neighbor search by quantizing each embedding and using L2 distance \cite{wu2017multiscale,guo2016quantization}, selecting the $k_v = 2$ closest images in embedding space.

We also use the iNaturalist scientific taxonomy $\mathcal{T(D)}$ to sample images at varying levels of taxonomic distance from $i_1$.
We select $k_t = 10$ taxonomically branched images by sampling two images each from the same \textsc{species} ($\ell=1$), \textsc{genus}, \textsc{family}, \textsc{order}, and \textsc{class} ($\ell = 5$) as $c$.
This yields 4,860 raw image pairs $(i_1, i_2)$.

\subsection{Language Collection}
For each image pair $(i_1, i_2)$, we elicit five natural language paragraphs describing the differences between them.

An annotator is instructed to write a paragraph (usually 2--5 sentences) comparing and contrasting the animal appearing in each image.
We instruct annotators not to explicitly mention the species (e.g., \textit{``Animal 1 is a penguin''}), and to instead focus on visual details (e.g., \textit{``Animal 1 has a black body and a white belly''}).
They are additionally instructed to avoid mentioning aspects of the background, scenery, or pose captured in the photograph (e.g., \textit{``Animal 2 is perched on a coconut''}).

We discard all annotations for an image pair where either image did not have at least $\frac{4}{5}$ positive ratings of image clarity.
This yields a total of 3,347 image pairs, annotated with 16,067 paragraphs.
Detailed statistics of the \datasetname~dataset are shown in Figure~\ref{figure:dataset}, and examples are provided in Figure~\ref{fig:examples}.
Further details of our both our algorithmic approach and dataset construction are given in Appendices \ref{sec:appendix-dataset-algo} and \ref{sec:appendix-dataset-details}.
\section{\modelname~Model}

\begin{figure*}[t!]

\begin{center}
\includegraphics[width=0.99\linewidth]{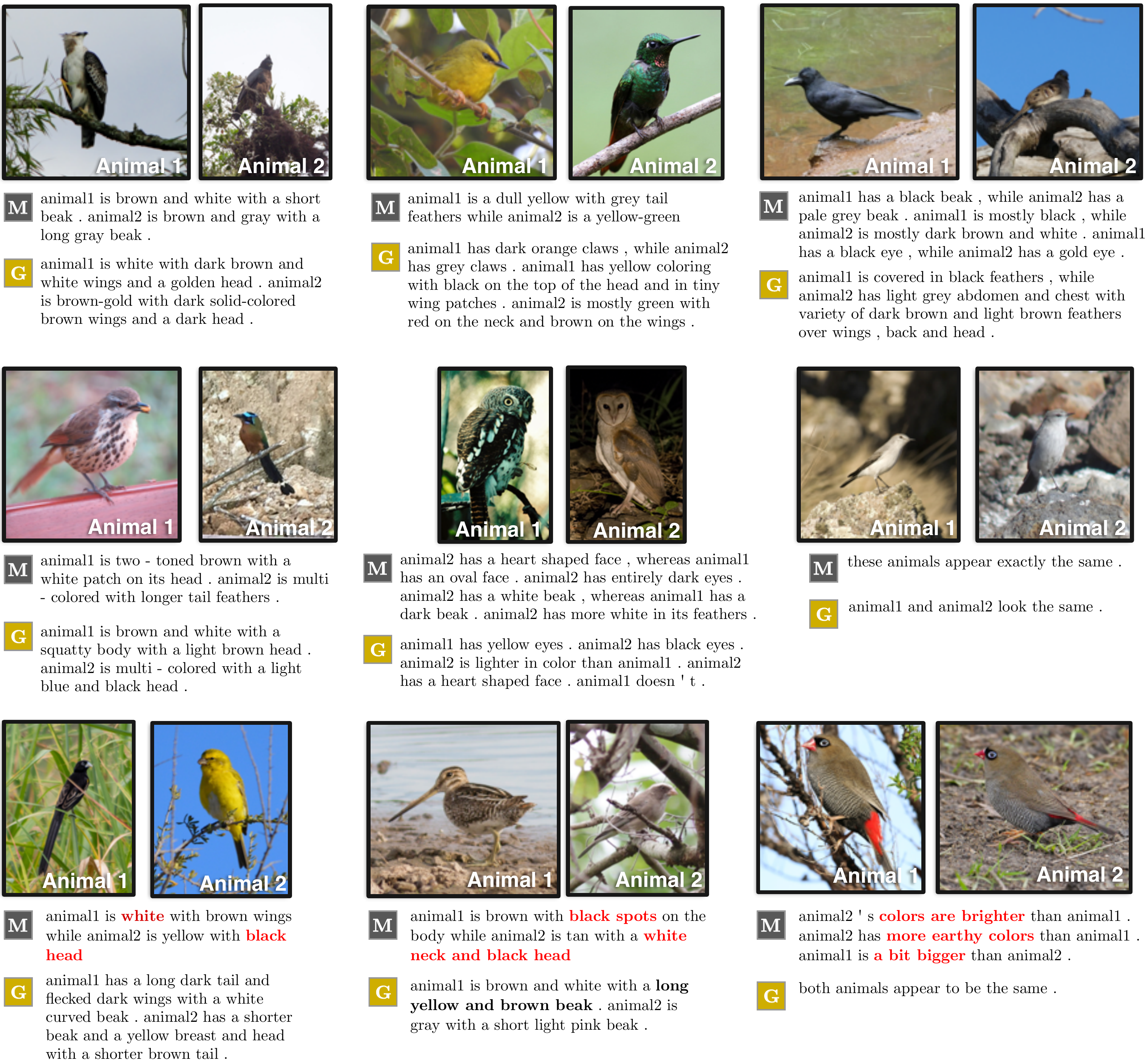}
\end{center}

\caption{Samples from the dev split of the proposed \datasetname~dataset, along with \modelname~model output (M) and one of five ground truth paragraphs (G). The second row highlights failure cases in red. The model produces \textit{coherent} descriptions of \textit{variable granularity}, though \textit{emphasis} and \textit{assignment} can be improved.
}
\label{fig:examples}

\end{figure*}

\paragraph{Task} Given two images $(i_1,i_2)$ as input, our task is to generate a natural language paragraph $t = x_1 \ldots x_n$ that compares the two images.

\paragraph{Architecture} Recent image captioning approaches \cite{xu2015show,sharma2018conceptual} extract image features using a convolutional neural network (CNN) which serve as input to a language decoder, typically a recurrent neural network (RNN) \cite{mikolov2010recurrent} or Transformer \cite{vaswani2017attention}.
We extend this paradigm with a joint encoding step and comparative module to study how best to encode and transform multiple latent image embeddings.
A schematic of the model is outlined in Figure~\ref{fig:model}, and its key components are described in the upcoming sections.

\subsection{Image Embedding}
Both input images are first processed using CNNs with shared weights.
In this work, we consider ResNet \cite{he2016deep} and Inception \cite{szegedy2017inception} architectures.
In both cases, we extract the representation from the deepest layer immediately before the classification layer.
This yields a dense 2D grid of local image feature vectors, shaped $(d,d,f)$.
We then flatten each feature grid into a $(d^2, f)$ shaped matrix:

\begin{align*}
    \mathbf{E}^1 &= \langle \mathbf{e}_{1,1}^1, \dots, \mathbf{e}_{d,d}^1 \rangle = \textsc{CNN}(i_1)\\
    \mathbf{E}^2 &= \langle \mathbf{e}_{1,1}^2, \dots, \mathbf{e}_{d,d}^2 \rangle = \textsc{CNN}(i_2)\\
\end{align*}

\subsection{Joint Encoding}

We define a joint encoding $\mathbf{J}$ of the images which contains both embedded images $(\mathbf{E}^1, \mathbf{E}^2)$, a mutated combination $(\mathbf{M})$, or both. We consider as possible mutations $\mathbf{M} \in \{ \mathbf{E}^1 + \mathbf{E}^2, \mathbf{E}^1 - \mathbf{E}^2, \max(\mathbf{E}^1, \mathbf{E}^2), \mathbf{E}^1 \odot \mathbf{E}^2\}$.
We try these encoding variants to explore whether simple mutations can effectively combine the image representations.

\subsection{Comparative Module}

Given the joint encoding of the images ($\mathbf{J}$), we would like to represent the differences in feature space ($\mathbf{C}$) in order to generate comparative descriptions. We explore two variants at this stage. The first is a direct passthrough of the joint encoding ($\mathbf{C} = \mathbf{J}$). This is analogous to ``standard'' CNN+LSTM architectures, which embed images and pass them directly to an LSTM for decoding. Because we try different joint encodings, a passthrough here also allows us to study their effects in isolation.

Our second variant is an $N$-layer Transformer encoder. This provides an additional self-attentive mutations over the latent representations $\mathbf{J}$.
Each layer contains a multi-headed attention mechanism ($\textsc{attn}_\textsc{mh}$).
The intent is that self-attention in Transformer encoder layers will guide comparisons across the joint image encoding.

Denoting \textsc{ln} as \textit{Layer Norm} and \textsc{ff} as \textit{Feed Forward}, with $\mathbf{C}_i$ as the output of the $i$th layer of the Transformer encoder, $\mathbf{C}_0 = \mathbf{J}$, and $\mathbf{C} = \mathbf{C}_N$:

\begin{align*}
    \mathbf{C}_i^H &= \textsc{ln}(\mathbf{C}_{i-1} + \textsc{attn}_\textsc{mh}(\mathbf{C}_{i-1})) \\
    \mathbf{C}_i &= \textsc{ln}(\mathbf{C}_i^H + \textsc{ff}(\mathbf{C}_i^H))
\end{align*}

\subsection{Decoder}

We use an $N$-layer Transformer decoder architecture to produce distributions over output tokens.
The Transformer decoder is similar to an encoder, but it contains an intermediary multi-headed attention which has access to the encoder's output $\mathbf{C}$ at every time step.

\begin{align*}
    \mathbf{D}_i^{H_1} & = \textsc{ln}(\mathbf{X} + \textsc{attn}_\textsc{mask,mh}(\mathbf{X})) \\
    \mathbf{D}_i^{H_2} & = \textsc{ln}(\mathbf{D}_i^{H_1} + \textsc{attn}_\textsc{mh}(\mathbf{D}_i^{H_1}, \mathbf{C})) \\
    \mathbf{D}_i & = \textsc{ln}(\mathbf{D}_i^{H_2} + \textsc{ff}(\mathbf{D}_i^{H_2})) \\    
\end{align*}

\noindent
Here we denote the text observed during training as $\mathbf{X}$, which is modulated with a position-based encoding and masked in the first multi-headed attention.

\section{Experiments}

\begin{table*}[]
\resizebox{\textwidth}{!}{%
\begin{tabular}{@{}llllllll@{}}
\toprule
 & \multicolumn{3}{c}{Dev} &  & \multicolumn{3}{c}{Test} \\ \cmidrule(lr){2-4} \cmidrule(l){6-8} 
 & BLEU-4 & ROUGE-L & CIDEr-D &  & BLEU-4 & ROUGE-L & CIDEr-D \\ \midrule
Most Frequent & 0.20 & 0.31 & \textbf{0.42} &  & 0.20 & 0.30 & \textbf{0.43} \\
Text-Only & 0.14 & 0.36 & 0.05 &  & 0.14 & 0.36 & 0.07 \\
Nearest Neighbor & 0.18 & 0.40 & 0.15 &  & 0.14 & 0.36 & 0.06 \\ \midrule
CNN + LSTM \cite{vinyals2015show} & 0.22 & 0.40 & 0.13 &  & 0.20 & 0.37 & 0.07 \\
CNN + Attn. + LSTM \cite{xu2015show} & 0.21 & 0.40 & 0.14 &  & 0.19 & 0.38 & 0.11 \\ \midrule
\modelname~-- Simple Joint Encoding & 0.23 & 0.44 & 0.23 &  & - & - & - \\
\modelname~-- No Comparative Module & 0.09 & 0.27 & 0.09 &  & - & - & - \\
\modelname~-- Small Decoder & 0.22 & 0.42 & 0.25 &  & - & - & - \\
\modelname~-- Full & \textbf{0.24} & \textbf{0.46} & 0.28 &  & \textbf{0.22} & \textbf{0.43} & 0.25 \\ \midrule
Human & 0.26 +/- 0.02 & 0.47 +/- 0.01 & 0.39 +/- 0.04 &  & 0.27 +/- 0.01 & 0.47 +/- 0.01 & 0.42 +/- 0.03 \\ \bottomrule
\end{tabular}
}
\caption{
Experimental results for comparative paragraph generation on the proposed dataset. For human captions, mean and standard deviation are given for a one-vs-rest scheme across twenty-five runs.
We observed that CIDEr-D scores had little correlation with description quality.
The \modelname~model benefits from a strong joint encoding and Transformer-based comparative module, achieving the highest BLEU-4 and ROUGE-L scores. 
}
\label{table:results}
\end{table*}

We train the \modelname~model to produce descriptions of the differences between images in the \datasetname~dataset.
We partition the dataset into train (80\%), val (10\%), and test (10\%) sections by splitting based on the pivot images $i_1$.
This ensures $i_1$ species are unique across the different splits.

We provide model hyperparameters and optimization details in Appendix ~\ref{sec:appendix-model-details}.

\subsection{Baselines and Variants}

The \textit{most frequent} paragraph baseline produces only the most observed description in the training data, which is that the two animals appear to be exactly the same.
\textit{Text-Only} samples captions from the training data according to their empirical distribution.
\textit{Nearest Neighbor} embeds both images and computes the lowest total $L_2$ distance to a training set pair, sampling a caption from it.
We include two standard neural baselines, CNN (+ Attention) + LSTM, which concatenate the images embeddings, optionally perform attention, and decode with an LSTM.
The main model variants we consider are a simple joint encoding ($\mathbf{J} = \langle \mathbf{E}^1, \mathbf{E}^2 \rangle$), no comparative module ($\mathbf{C} = \mathbf{J}$), a small (1-layer) decoder, and our full \modelname~model. We also try several other ablations and model variants, which we describe later.

\subsection{Quantitative Results}

\paragraph{Automatic Metrics} We evaluate our model using three machine-graded text metrics: BLEU-4 \cite{papineni2002bleu}, ROUGE-L \cite{lin2004rouge}, and CIDEr-D \cite{vedantam2015cider}.
Each generated paragraph is compared to all five reference paragraphs.

For human performance, we use a one-vs-rest scheme to hold one reference paragraph out and compute its metric using the other four.
We average this score across twenty-five runs over the entire split in question.

Results using these metrics are given in Table~\ref{table:results} for the main baselines and model variants.
We observe improvement across BLEU-4 and ROUGE-L scores compared to baselines.
Curiously, we observe that the CIDEr-D metric is susceptible to common patterns in the data; our model, when stopped at its highest CIDEr-D score, outputs a variant of, \textit{``these animals appear exactly the same''} for 95\% of paragraphs, nearly mimicking the behavior of the most frequent paragraph (\textit{Freq.}) baseline.
The corpus-level behavior of CIDEr-D gives these outputs a higher score.
We observed anecdotally higher quality outputs correlated with ROUGE-L score, which we verify using a human evaluation (paragraph after next).

\paragraph{Ablations and Model Variants} We ablate and vary each of the main model components, running the automatic metrics to study coarse changes in the model's behavior.
Results for these experiments are given in Table~\ref{table:fullvariants}.
For the \emph{joint encoding}, we try combinations of four element-wise operations with and without both encoded images.
To study the \emph{comparative module} in greater detail, we examine its effect on the top three joint encodings: $(i_1, i_2, -)$, $-$, and $\odot$.
After fixing the best joint encoding and comparative module, we also try variations of the \emph{decoder} (Transformer depth), as well as \emph{decoding algorithms} (greedy decoding, multinomial sampling, and beamsearch).

Overall, we we see that the choice of joint encoding requires a balance with the choice of comparative module. More disruptive joint encodings (like element-wise multiplication $\odot$) appear too destructive when passed directly to a decoder, but yield the best performance when paired with a deep comparative module. Others (like subtraction) function moderately well on their own, and are further improved when a comparative module is introduced.

\begin{table*}[ht]
\resizebox{\textwidth}{!}{%
\begin{tabular}{@{}llllllccllll@{}}
\toprule
\multicolumn{6}{c}{Joint Encoding} & \multicolumn{1}{l}{} & \multicolumn{1}{l}{} & \multirow{2}{*}[-0.5em]{\begin{tabular}[c]{@{}c@{}}Decoding\\Algorithm\end{tabular}}  & \multicolumn{3}{c}{Dev} \\ \cmidrule(r){1-6} \cmidrule(l){10-12} 
$i_1$ & $i_2$ & $-$ & $+$ & max & $\odot$ & \multicolumn{1}{l}{Comparative Module} & Decoder & & BLEU-4 & ROUGE-L & CIDEr-D \\ \midrule
\hlcell \textbf{\checkmark} & \hlcell \textbf{\checkmark} & \hlcell \textbf{} & \hlcell \textbf{} & \hlcell \textbf{} & \hlcell \textbf{} & \multirow{10}{*}{\begin{tabular}[c]{@{}c@{}}6-Layer\\ Transformer\end{tabular}} & \multirow{10}{*}{\begin{tabular}[c]{@{}c@{}}6-Layer\\ Transformer\end{tabular}} & \multirow{10}{*}{Beamsearch} & 0.23 & 0.44 & 0.23 \\
\hlcell \textbf{} & \hlcell \textbf{} & \hlcell \textbf{\checkmark} & \hlcell \textbf{} & \hlcell \textbf{} & \hlcell \textbf{} &  &  &  & 0.23 & 0.45 & 0.27 \\
\hlcell \textbf{} & \hlcell \textbf{} & \hlcell \textbf{} & \hlcell \textbf{\checkmark} & \hlcell \textbf{} & \hlcell \textbf{} &  &  &  & \textbf{0.24} & 0.43 & \textbf{0.28} \\
\hlcell \textbf{} & \hlcell \textbf{} & \hlcell \textbf{} & \hlcell \textbf{} & \hlcell \textbf{\checkmark} & \hlcell \textbf{} &  &  &  & 0.23 & 0.43 & 0.24 \\
\hlcell \textbf{} & \hlcell \textbf{} & \hlcell \textbf{} & \hlcell \textbf{} & \hlcell \textbf{} & \hlcell \textbf{\checkmark} &  &  &  & \textbf{0.24} & \textbf{0.46} & \textbf{0.28} \\
\hlcell \textbf{\checkmark} & \hlcell \textbf{\checkmark} & \hlcell \textbf{\checkmark} & \hlcell \textbf{} & \hlcell \textbf{} & \hlcell \textbf{} &  &  &  & 0.22 & 0.44 & 0.22 \\
\hlcell \textbf{\checkmark} & \hlcell \textbf{\checkmark} & \hlcell \textbf{} & \hlcell \textbf{\checkmark} & \hlcell \textbf{} & \hlcell \textbf{} &  &  &  & 0.22 & 0.42 & 0.25 \\
\hlcell \textbf{\checkmark} & \hlcell \textbf{\checkmark} & \hlcell \textbf{} & \hlcell \textbf{} & \hlcell \textbf{\checkmark} & \hlcell \textbf{} &  &  &  & 0.21 & 0.42 & 0.22 \\
\hlcell \textbf{\checkmark} & \hlcell \textbf{\checkmark} & \hlcell \textbf{} & \hlcell \textbf{} & \hlcell \textbf{} & \hlcell \textbf{\checkmark} &  &  &  & 0.22 & 0.43 & 0.23 \\
\hlcell \textbf{\checkmark} & \hlcell \textbf{\checkmark} & \hlcell \textbf{\checkmark} & \hlcell \textbf{\checkmark} & \hlcell \textbf{\checkmark} & \hlcell \textbf{\checkmark} &  &  &  & 0.21 & 0.43 & 0.20 \\ \midrule
 &  &  &  &  & \checkmark & \multicolumn{1}{l}{\hlcompcell Passthrough} & \multirow{4}{*}{\begin{tabular}[c]{@{}c@{}}6-Layer\\ Transformer\end{tabular}} & \multicolumn{1}{c}{\multirow{4}{*}{Beamsearch}} & 0.00 & 0.02 & 0.00 \\
 &  &  &  &  & \checkmark & \multicolumn{1}{l}{\hlcompcell 1-L Transformer} &  & \multicolumn{1}{c}{} & \textbf{0.24} & 0.44 & 0.27 \\
 &  &  &  &  & \checkmark & \multicolumn{1}{l}{\hlcompcell 3-L Transformer} &  & \multicolumn{1}{c}{} & \textbf{0.24} & 0.44 & 0.27 \\
 &  &  &  &  & \checkmark & \multicolumn{1}{l}{\hlcompcell 6-L Transformer} &  & \multicolumn{1}{c}{} & \textbf{0.24} & \textbf{0.46} & \textbf{0.28} \\ \midrule[0.03em]
&  & \checkmark &  &  &  & \multicolumn{1}{l}{\hlcompcell Passthrough} & \multirow{4}{*}{\begin{tabular}[c]{@{}c@{}}6-Layer\\ Transformer\end{tabular}} & \multicolumn{1}{c}{\multirow{4}{*}{Beamsearch}} & 0.22 & 0.40 & 0.22 \\
&  & \checkmark &  &  &  & \multicolumn{1}{l}{\hlcompcell 1-L Transformer} &  & \multicolumn{1}{c}{} & 0.21 & 0.41 & 0.26 \\
&  & \checkmark &  &  &  & \multicolumn{1}{l}{\hlcompcell 3-L Transformer} &  & \multicolumn{1}{c}{} & 0.22 & 0.41 & 0.22 \\
&  & \checkmark &  &  &  & \multicolumn{1}{l}{\hlcompcell 6-L Transformer} &  & \multicolumn{1}{c}{} & \textbf{0.23} & \textbf{0.45} & \textbf{0.27} \\ \midrule[0.03em]
\checkmark & \checkmark & \checkmark &  &  &  & \multicolumn{1}{l}{\hlcompcell Passthrough} & \multirow{4}{*}{\begin{tabular}[c]{@{}c@{}}6-Layer\\ Transformer\end{tabular}} & \multicolumn{1}{c}{\multirow{4}{*}{Beamsearch}} & 0.09 & 0.27 & 0.09 \\
\checkmark & \checkmark & \checkmark &  &  &  & \multicolumn{1}{l}{\hlcompcell 1-L Transformer} &  & \multicolumn{1}{c}{} & \textbf{0.24} & 0.43 & 0.24 \\
\checkmark & \checkmark & \checkmark &  &  &  & \multicolumn{1}{l}{\hlcompcell 3-L Transformer} &  & \multicolumn{1}{c}{} & 0.22 & 0.42 & \textbf{0.26} \\
\checkmark & \checkmark & \checkmark &  &  &  & \multicolumn{1}{l}{\hlcompcell 6-L Transformer} &  & \multicolumn{1}{c}{} & 0.22 & \textbf{0.44} & 0.22 \\
\midrule
 &  &  &  &  & \checkmark & \multirow{3}{*}{\begin{tabular}[c]{@{}c@{}}6-Layer\\ Transformer\end{tabular}} & \multicolumn{1}{l}{\hldecodecell 1-L Transformer} & \multicolumn{1}{c}{\multirow{3}{*}{Beamsearch}} & 0.22 & 0.42 & 0.25 \\
 &  &  &  &  & \checkmark &  & \multicolumn{1}{l}{\hldecodecell 3-L Transformer} & \multicolumn{1}{c}{} & 0.23 & 0.42 & 0.25 \\
 &  &  &  &  & \checkmark &  & \multicolumn{1}{l}{\hldecodecell 6-L Transformer} & \multicolumn{1}{c}{} & \textbf{0.24} & \textbf{0.46} & \textbf{0.28} \\ \midrule
 &  &  &  &  & \checkmark & \multirow{3}{*}{\begin{tabular}[c]{@{}c@{}}6-Layer\\ Transformer\end{tabular}} & \multicolumn{1}{l}{\multirow{3}{*}{\begin{tabular}[c]{@{}c@{}}6-Layer\\ Transformer\end{tabular}}} & \hlalgocell Greedy & 0.21 & 0.44 & 0.18 \\
 &  &  &  &  & \checkmark &  & \multicolumn{1}{l}{} & \hlalgocell Multinomial & 0.20 & 0.42 & 0.16 \\
 &  &  &  &  & \checkmark &  & \multicolumn{1}{l}{} & \hlalgocell Beamsearch & \textbf{0.24} & \textbf{0.46} & \textbf{0.28} \\
\bottomrule
\end{tabular}
}
\caption{Variants and ablations for the \modelname~model. We find the best performing combination is an elementwise multiplication ($\odot$) for the joint encoding, a 6-layer Transformer comparative module, a 6-layer Transformer decoder, and using beamsearch to perform inference.}
\label{table:fullvariants}
\end{table*}

\paragraph{Human Evaluation} To verify our observations about model quality and automatic metrics, we also perform a human evaluation of the generated paragraphs.
We sample 120 instances from the test set, taking twenty each from the six categories for choosing comparative images (visual similarity in embedding space, plus five taxonomic distances).
We provide annotators with the two images in a random order, along with the output from the model at hand.
Annotators must decide which image contains \textit{Animal 1}, and which contains \textit{Animal 2}, or they may say that there is no way to tell (e.g., for a description like \textit{``both look exactly the same''}).

We collect three annotations per datum, and score a decision only if $\geq$ 2/3 annotators made that choice.
A model receives +1 point if annotators decide correctly, 0 if they cannot decide or agree there is no way to tell, and -1 point if they decide incorrectly (label the images backwards).
This scheme penalizes a model for confidently writing incorrect descriptions.
The total score is then normalized to the range $[-1, 1]$.
Note that \textit{Human} uses one of the five gold paragraphs sampled at random.

Results for this experiment are shown in Table~\ref{table:humaneval}.
In this measure, we see the frequency and text-only baselines now fall flat, as expected.
The frequency baseline never receives any points, and the text-only baseline is often penalized for incorrectly guessing.
Our model is successful at making distinctions between visually distinct species (\textsc{Genus} column and ones further right), which is near the challenge level of current fine-grained visual classification tasks.
However, it struggles on the two data subsets with highest visual similarity (\textsc{Visual, Species}).
The significant gap between all methods and human performance in these columns indicates ultra fine-grained distinctions are still possible for humans to describe, but pose a challenge for current models to capture.

\subsection{Qualitative Analysis}

In Figure~\ref{fig:examples}, we present several examples of the model output for pairs of images in the dev set, along with one of the five reference paragraphs.
In the following section, we split an analysis of the model into two parts: largely positive findings, as well as common error cases.

\subsubsection*{Positive Findings}

We find that the model exhibits \textbf{dynamic granularity}, by which we mean that it adjusts the magnitude of the descriptions based on the scale of differences between the two animals.
If two animals are quite similar, it generates fine-grained descriptions such as, \textit{``Animal 2 has a slightly more curved beak than Animal 1,''} or \textit{``Animal 1 is more iridescent than Animal 2.''}
If instead the two animals are very different, it will generate text describing larger-scale differences, like, \textit{``Animal 1 has a much longer neck than Animal 2,''} or \textit{``Animal 1 is mostly white with a black head. Animal 2 is almost completely yellow.''} 

We also observe that the model is able to produce coherent paragraphs of \textbf{varying linguistic structure.}
These include a range of comparisons set up across both single and multiple sentences.
For example, one it generates straightforward comparisons of the form, \textit{Animal 1 has X, while Animal 2 has Y.}
But it also generates contrastive expressions with longer dependencies, such as \textit{Animal 1 is X, Y, and Z. Animal 2 is very similar, except W.}
Furthermore, the model will mix and match different comparative structures within a single paragraph.

Finally, in addition to varying linguistic structure, we find the model is able to produce \textbf{coherent semantics} through a series of statements. For example, consider the following full output: \textit{``Animal 1 has a very long neck compared to Animal 2. Animal 1 has shorter legs than Animal 2. Animal 1 has a black beak, Animal 2 has a brown beak. Animal 1 has a yellow belly. Animal 2 has darker wings than Animal 1.''} The range of concepts in the output covers \textit{neck, legs, beak, belly, wings} without repeating any topic or getting sidetracked.

\begin{table}[]
\resizebox{0.99\linewidth}{!}{%
\small
\begin{tabular}{@{}lrrrrrr@{}}
\toprule
 & \textsc{Visual} & \textsc{Species} & \textsc{Genus} & \textsc{Family} & \textsc{Order} & \textsc{Class} \\ \midrule
Freq. & 0.00 & 0.00 & 0.00 & 0.00 & 0.00 & 0.00 \\
Text-Only & 0.00 & -0.10 & -0.05 & 0.00 & 0.15 & -0.15 \\
CNN + LSTM & -0.15 & \textbf{0.20} & 0.15 & \textbf{0.50} & 0.40 & 0.15 \\
CNN + Attn. + LSTM & \textbf{0.15} & 0.15 & 0.15 & -0.05 & 0.05 & 0.20 \\
\modelname & 0.10 & -0.10 & \textbf{0.35} & 0.40 & \textbf{0.45} & \textbf{0.55} \\ \midrule
Human & 0.55 & 0.55 & 0.85 & 1.00 & 1.00 & 1.00 \\ \bottomrule
\end{tabular}
}
\caption{Human evaluation results on 120 test set samples, twenty per column. Scale: -1 (perfectly wrong) to 1 (perfectly correct). Columns are ordered left-to-right by increasing distance.
Our model outperforms baselines for several distances, though highly similar comparisons still prove difficult.
}
\label{table:humaneval}
\end{table}

\subsubsection*{Error Analysis}

We also observe several patterns in the model's shortcomings.
The most prominent error case is that the model will sometimes \textbf{hallucinate differences} (Figure~\ref{fig:examples}, bottom row).
These range from pointing out significant changes that are missing (e.g., \textit{``a black head''} where there is none (Fig.~\ref{fig:examples}, bottom left)), to clawing at subtle distinctions where there are none (e.g., \textit{``[its] colors are brighter \ldots and [it] is a bit bigger''} (Fig.~\ref{fig:examples}, bottom right)).
We suspect that the model has learned some associations between common features in animals, and will sometimes favor these associations over visual evidence.

The second common error case is \textbf{missing obvious distinctions}.
This is observed in Fig.~\ref{fig:examples} (bottom middle), where the prominent beak of Animal 1 is ignored by the model in favor of mundane details.
While outlying features make for lively descriptions, we hypothesize that the model may sometimes avoid taking them into account given its per-token cross entropy learning objective.

Finally, we also observe the model sometimes \textbf{swaps which features are attributed to which animal}.
This is partially observed in Fig.~\ref{fig:examples} (bottom left), where the \textit{``black head''} actually belongs to Animal 1, not Animal 2.
We suspect that mixing up references may be a trade-off for the representational power of attending over both images; there is no explicit bookkeeping mechanism to enforce which phrases refer to which feature comparisons in each image.
\section{Related Work}

Employing visual comparisons to elicit focused natural language observations was proposed by \citet{maji2012discovering}. \citet{zou2015crowdsourcing} studied this tactic in the context of crowdsourcing, and \citet{su2017reasoning} performed a large scale investigation in the aircraft domain, using reference games to evoke attribute phrases.
We take inspiration from these works.

Previous work has collected natural language captions of bird photographs: CUB Captions \cite{reed2016learning} and CUB-Justify \cite{vedantam2017context} are both language annotations on top of the CUB-2011 dataset of bird photographs \cite{wah2011caltech}.
In addition to describing two photos instead of one, the language in our dataset is more complex by comparison, containing a diversity of comparative structures and implied semantics.
We also collect our data without an anatomical guide for annotators, yielding everyday language in place of scientific terminology.

Conceptually, our paper offers a complementary approach to works that generate single-image, class or image-discriminative captions \cite{hendricks2016generating,vedantam2017context}.
Rather than discriminative captioning, we focus on comparative language as a means for bridging the gap between varying granularities of visual diversity.

Methodologically, our work is most closely related to the Spot-the-diff dataset \cite{jhamtani2018learning} and other recent work on change captioning \cite{park2019robust,tan2019expressing}.
While change captioning compares two images with few changing pixels (e.g., surveillance footage), we consider image pairs with no pixel overlap, motivating our stratified sampling approach to select comparisons.

Finally, the recently released NLVR$^2$ dataset \cite{suhr2018corpus} introduces a challenging natural language reasoning task using two images as context. Our work instead focuses on generating comparative language rather than reasoning.
\section{Conclusion}

We present the new \datasetname~dataset and \modelname~model for generating comparative explanations of fine-grained visual differences.
This dataset features paragraph-length, adaptively detailed descriptions written in everyday language.
We hope that continued study of this area will produce models that can aid humans in critical domains like citizen science.

\bibliography{main}
\bibliographystyle{acl_natbib}

\appendix
\section{Algorithmic Approach to Dataset Construction}
\label{sec:appendix-dataset-algo}

We present here an algorithmic approach to collecting a dataset of image pairs with natural language text describing their differences.
The central challenge is to balance empirical desiderata---mainly, sample coverage and model relevance---with practical constraints of data quality and cost.
This algorithmic approach underpins the dataset collection we outlined in the
paper body.

\subsection{Goals}

Our goal is to collect a dataset of tuples $(i_1, i_2, t)$, where $i_1$ and $i_2$ are images, and $t$ is a textual comparison of them.
We can consider each image $i$ as drawn from some domain $\mathcal{D} \in \{\textit{furniture, trees, ...}\}$, or a completely open domain of all concepts.
There are several criteria we would like to balance:

\begin{enumerate}
    \item \textbf{Coverage} A dataset should sufficiently cover $\mathcal{D}$ so that generalization across the space is possible.
    
    \item \textbf{Relevance} Given the capabilities for models to distinguish $i_1$ and $i_2$, $t$ should provide value.
    
    \item \textbf{Comparability} Each pair ($i_1$, $i_2$) must have sufficient structural similarities that a human annotator can reasonably write $t$ comparing them. Pairs that are too different will yield lengthy and uninteresting descriptions without direct contrasting statements. Pairs that are too similar for human perception may yield \textit{``I can't see any difference.''}\footnote{This hints at the same sweet spot the fine-grained visual classification (FGVC) community studies, like cars \cite{KrauseStarkDengFei-Fei_3DRR2013}, aircraft \cite{maji2013fine}, dogs \cite{KhoslaYaoJayadevaprakashFeiFei_FGVC2011}, and birds \cite{wah2011caltech,van2018inaturalist}.}
    
    \item \textbf{Efficiency} Image judgements and textual annotations require human labor. With a fixed budget, we would like to yield a dataset of the largest size possible.
\end{enumerate}

We describe sampling algorithms for addressing these issues given the choice of a domain. 

\subsection{Pivot-Branch Sampling}
Drawing a single image $i$ from domain $\mathcal{D}$, there is a chance $p \in [0, 1]$ that each image is ill-suited for comparisons. 
For example, $i$ might be out-of-focus or contain multiple instances.

If a pair of images is drawn, and each has probability $p$ of being discarded, then $\frac{1}{(1-p)^2}$ times more pairs must be selected and annotated.
For example, if $p = \frac{2}{3}$, then the annotation cost is scaled by 2.25. 
This severely impacts annotation \textit{efficiency}.

To combat this, we employ a stratified sampling strategy we call \textit{pivot-branch sampling}.
Each image on one side of the comparison (say, $i_\text{pivot}$) is vetted individually, and $k$ images on the other side (say, $i_\text{branch}$) are sampled to produce pairs. 
With $k$-times fewer $i_\text{pivot}$ images, it is feasible to check each instance for usability.
This lowers the annotation cost scale to $\frac{1}{1-p}$ (e.g., with $p = \frac{2}{3}$, this is 1.5).

Splitting our selection from $\mathcal{D}$ into two parts allows us to define two distinct sampling strategies.
One choice is for $s_\text{pivot}(\mathcal{D})$ to select pivot images.
The second is for $s_\text{branch}(\mathcal{D}, i_\text{pivot}, k)$ to sample $k$ images given a single pivot image.

\subsection{Designing $s_\text{pivot}(\mathcal{D})$}

Selecting $i_\text{pivot}$ are important because each will contribute to $k$ image pairs in a dataset.
Here we consider the case where there are class labels $c \in \mathcal{C}$ available for each image in the domain.
We propose selecting $s_\text{pivot}$ to sample uniformly over $\mathcal{C}$.
This strategy attempts to provide \text{coverage} over $\mathcal{D}$ using class labels as a coarse measure of diversity.
It accounts for category-level dataset bias (e.g., where most images belong to only a few classes).
This pushes the need to address \textit{relevance} and \textit{comparability} to the sampling procedure for branched images.

\subsection{Designing $s_\text{branch}(\mathcal{D}, i_\text{pivot}, k)$}
Given each pivot image $i_\text{pivot}$, we will choose $k$ images from $\mathcal{D}$ for comparison.
We can make use of additional functions and structure available on $\mathcal{D}$:

\begin{align*}
    & \mathcal{V}(i_1,i_2) \rightarrow [0, 1]\\[-0.5em]
    & \text{A function that measures the visual similarity} \\[-0.5em]
    & \text{between any two images.} \\[0.5em]
    & \mathcal{T}(\mathcal{D})\\[-0.5em]
    & \text{A taxonomy over }\mathcal{D}\text{, with image class labels}\\[-0.5em]
    & c \in \mathcal{C} \text{ as leaves.} \\
\end{align*}

We can partition $k = k_v + k_t$ to sample $k_v$ visually-similar images using and $k_t$ taxonomically related images.
A simple strategy for visually similar images is to pick

$$
\argmin_{i' \in \mathcal{D}, i' \neq i_\text{pivot}}\mathcal{V}(i_\text{pivot}, i')
$$

\noindent
$k_v$ times without replacement. This samples the $k_v$ most visually similar images to $i_\text{pivot}$, excluding the image itself.

To employ taxonomic information, we propose a walk over mutually exclusive subsets of $\mathcal{T}(\mathcal{D})$.
We define a function $a_\mathcal{T(D)}(c, \ell)$ that gives the set of other taxonomic leaves that share a common ancestor exactly $\ell$ taxonomic levels above $c$, and no levels lower. More formally, if we use $p(c,c',\ell)$ to express that $c$ and $c'$ share a parent $\ell$ taxonomic levels above $c$, then we can define:
$$
a_{\mathcal{T(D)}}(c, \ell) = \{c' : p(c,c',\ell) \land \nexists_{\ell' < \ell} \, p(c,c',\ell')\}
$$

The function $a_{\mathcal{T(D)}}(c, \ell)$ partitions the taxonomy $\mathcal{T(D)}$ into disjoint subtrees. For example, $a_{\mathcal{T(D)}}(c, 1)$ are the set of sibling classes to $c$ which share its direct parent; $a_{\mathcal{T(D)}}(c, 2)$ are the set of cousin classes to $c$ which share its grandparent, but \textit{not} its parent.

We can employ $a_{\mathcal{T(D)}}(c,\ell)$ by choosing class $c$ from our pivot image $i_\text{pivot}$ and varying $\ell$. As we increase $\ell$, we define mutually exclusive sets of classes with greater taxonomic distance from $c$. 

To sample images using this scheme, we can further split our $k_t$ budget for taxonomically sampled images into $k_t = k_{t_1} + k_{t_2} + \dots + k_{t_\ell}$ for $\ell$ different levels. Then, if we write the set of classes $\mathcal{C}_\ell = a_{\mathcal{T(D)}}(c,\ell)$, we can sample $k_{t_\ell}$ images from $\mathcal{C}$.
One scheme is to perform round-robin sampling: rotate through each class $c_\ell \in \mathcal{C}_\ell$ and sample sample one image from each until $k_{t_\ell}$ are chosen.

\subsection{Analyzing $s_\text{branch}(\mathcal{D}, i_\text{pivot}, k)$}

Given a good visual similarity function $\mathcal{V}$, image pairs will exhibit enough similarity to satisfy requirement that they be semantically close enough to be \textit{comparable}.
They may also be so visually similar that comparability is difficult. However, this aspect counter-balances with \textit{relevance}: if $\mathcal{V}(i_1,i_2)$ is small under a visual model, but their differences are describable by humans, their difference description has high value because it distinguishes two points with high similarity in visual embeddings space.

The use of the taxonomy $\mathcal{T(D)}$ complements $\mathcal{V}$ by providing controllable \textit{coverage} over $\mathcal{D}$ while maintaining \textit{relevance} and \textit{comparability}. Tuning the range of $\ell$ values used in the taxonomic splits $a_{\mathcal{T(D)}}(c, \ell)$ ensures comparability is maintained. Clamping $\ell$ below a threshold ensures images have sufficient similarity, and controlling the proportion of $k_{t_\ell}$ for small values of $\ell$ mitigates the risk of too-similar image pairs.

Similarly, we can adjust the relevance of taxonomic sampling by controlling the distribution of $k_{t_1} \dots k_{t_\ell}$ with respect to the particular structure of the taxonomy $\mathcal{T(D)}$.
If the taxonomy is well-balanced, then fixing a constant $k_{t_\ell}$ will draw proportionally more samples from subtrees close to $c$.
This can be seen by considering that $a_{\mathcal{T(D)}}(c, \ell)$ defines exponentially larger subsets of $\mathcal{T(D)}$ as $\ell$ increases.
Drawing the same number of samples from each subset biases the collection towards relevant pairs (which should be more difficult to distinguish) while maintaining sparse coverage over the entirety of $\mathcal{D}$.

\section{Details for Constructing \datasetname~Dataset}
\label{sec:appendix-dataset-details}

We provide here additional details for constructing the \datasetname~ dataset. This is meant to link the high level overview in Section~\ref{sec:dataset} with the algorithmic approach presented in the previous section (Appendix~\ref{sec:appendix-dataset-algo}).

\subsection{Clarity}

To build a dataset emphasizing fine-grained comparisons between two animals, we impose stricter restrictions on the images than iNaturalist research-grade observations (photographs). An iNaturalist observation that is research-grade indicates the community has reached consensus on the animal's species, that the photo was taken in the wild, and several other qualifications.\footnote{More details on iNaturalist research-grade specification: \url{https://www.inaturalist.org/pages/help\#quality}} We include four additional criteria that we define together as \textit{clarity}:

\begin{enumerate}

    \item \textbf{Single instance}: A photo must include only a single instance of the target species. Bird photography often includes flocks in trees, in the air, or on land. In addition, some birds appear in male/female pairs. For our dataset, all of those photos must be discarded.

    \item \textbf{Animal}: A photo must include the animal itself, rather than a record of it (e.g., tracks).
    
    \item \textbf{Focus}: A photo must be sufficiently in-focus to describe the animal in detail.

    \item \textbf{Visibility}: The animal in the photo must not be too obscured by the environment, and must take up enough pixels in the photo to be clearly described.

\end{enumerate}

\subsection{Pivot Images}

To pick pivot images, we first uniformly sample from the set of 9k species in the taxonomic \textsc{class} \textit{Aves} in iNaturalist.
We consider only species with at least four recorded observations to promote the likelihood that at least one image is \textit{clear}.
We also perform look-ahead branch sampling to ensure that a species will yield sufficient comparisons taxonomically.
For each species, we manually review four images sampled from this species to select the clearest image to use as the pivot image.
If none are suitable, we move to the next species.
With this manual process, we select 405 species and corresponding photographs to use as pivot $i_1$ images.

\subsection{Branching Images}

See Section~\ref{sec:dataset-branching-images} for the description of selecting $k_v=2$ visually similar branching images using a function $\mathcal{V}(i_1, i_2)$. We highlight here the use of the taxonomy $\mathcal{T(D)}$ to select $k_t=10$ branching images with varying levels of taxonomic distance.

For the class $c$ corresponding to image $i_1$, we split the taxonomic tree into \emph{disjoint} subtrees rooted $\ell \in \{1..5\}$ taxonomic levels above $c$.
Each higher level \emph{excludes} the levels beneath it.
For example, at $\ell = 1$ we consider all images of the same species as $i_1$; at $\ell = 2$, we consider all images of the same genus as $i_1$, but that have a \textit{different} species.
We set each $k_{t_\ell} = 2$ for a total of $k_t = 10$.

\subsection{Annotations}

\paragraph{Clarity} Annotators first label whether $i_1$ and $i_2$ are \emph{clear}.
While we manually verified each $i_1$ is clear, each $i_2$ must still be vetted.\footnote{Annotators would occasionally agree that a particular $i_1$ images was in fact unclear, upon which we removed it and all corresponding pairs from the dataset.}
Starting from 405 pivot images $i_1$, and selecting $k=12$ branching images $i_2$ for each, we annotated a total of 4,860 image pairs.
After restricting images to have $\geq \frac{4}{5}$ positive clarity judgments, we ended up with the 3,347 image pairs in our dataset, a retention rate of 68.9\%.

\paragraph{Quality} We vet each annotator individually by manually reviewing five reference annotations from a pilot round, and perform random quality assessments during data collection.
We found that manually vetting the writing quality and guideline adherence of each individual annotator vital for ensuring high data quality.

\section{Model Details}
\label{sec:appendix-model-details}
For the image embedding component of our model, we use a ResNet-101 network as our CNN.
We use a model pretrained on ImageNet and fix the CNN weights before starting training for our task.
We also experimented with an Inception-v4 model, but found ResNet-101 to have better performance.

For both the Transformer encoder and decoder, we use $N=6$ layers, a hidden size of 512, 8 attention heads, and dot product self-attention.
Each paragraphs is clipped at 64 tokens during training (chosen empirically to cover 94\% of paragraphs).
The text is preprocessed using standard techniques (tokenization, lowercasing), and we replace mentions referring to each image with special tokens \textsc{animal1} and \textsc{animal2}.

For inference, we experiment with greedy decoding, multinomial sampling, and beam search.
Beam search performs best, so we use it with a beam size of 5 for all reported results (except the decoding ablations, where we report each).

We train with Adagrad for 700k steps using a learning rate of .01 and batch size of 2048.
We decay the learning rate after 20k steps by a factor of 0.9.
Gradients are clipped at a magnitude of 5.

\section{Image Attributions}

The table above provides attributions for all photographs used in this paper.

\begin{table}[]
\resizebox{0.99\textwidth}{!}{%
\small
\begin{tabular}{@{}ll@{}}
\toprule
Photograph & Attribution \\ \midrule
Fig. 1: Top and bottom left & salticidude \textit{(CC BY-NC 4.0)} \minitab \scriptsize{\url{https://www.inaturalist.org/observations/20863620}} \\
Fig. 1: Top right & Patricia Simpson \textit{(CC BY-NC 4.0)} \minitab \scriptsize{\url{https://www.inaturalist.org/observations/1032161}} \\
Fig. 1: Bottom right & kalamurphyking \textit{(CC BY-NC-ND 4.0)} \minitab \scriptsize{\url{https://www.inaturalist.org/observations/9376125}} \\ \midrule
Fig. 2: Top left & Ryan Schain \minitab \scriptsize{\url{https://macaulaylibrary.org/asset/58977041}} \\
Fig. 2: Top right & Anonymous eBirder \minitab \scriptsize{\url{https://www.allaboutbirds.org/guide/Song_Sparrow/media-browser/66116721}} \\
Fig. 2: Right, 2nd from top & Garth McElroy/VIREO \minitab \scriptsize{\url{https://www.audubon.org/field-guide/bird/song-sparrow\#photo3}} \\
Fig. 2: Right, 3rd from top & Myron Tay \minitab \scriptsize{\url{http://orientalbirdimages.org/search.php?Bird_ID=2104&Bird_Image_ID=61509&p=73}} \\
Fig. 2: Right, 4th from top & Brian Kushner \minitab \scriptsize{\url{https://www.audubon.org/field-guide/bird/blue-jay}} \\
Fig. 2: Bottom, left & A. {\cyrrm{{SHCH}erbakov}} \\
Fig. 2: Bottom, right & prepa3tgz-11bwv518 \textit{(CC BY-NC 4.0)} \minitab \scriptsize{\url{https://www.inaturalist.org/observations/23184228}} \\ \midrule
Fig. 4: Top & jmaley \textit{(CC0 1.0)} \minitab \scriptsize{\url{https://www.inaturalist.org/observations/31619615}} \\
Fig. 4: Bottom & lorospericos \textit{(CC BY-NC 4.0)} \minitab \scriptsize{\url{https://www.inaturalist.org/observations/30605775}} \\ \midrule
Fig. 5: Top left, left & wildlife-naturalists \textit{(CC BY-NC 4.0)} \minitab \scriptsize{\url{https://www.inaturalist.org/photos/13223248}} \\
Fig. 5: Top left, right & Colin Barrows \textit{(CC BY-NC-SA 4.0)} \minitab \scriptsize{\url{https://www.inaturalist.org/photos/2642277}} \\
Fig. 5: Top middle, left & charley \textit{(CC BY-NC 4.0)} \minitab \scriptsize{\url{https://www.inaturalist.org/photos/13379419}} \\
Fig. 5: Top middle, right & guyincognito \textit{(CC BY-NC 4.0)} \minitab \scriptsize{\url{https://www.inaturalist.org/photos/26314681}} \\
Fig. 5: Top right, left & Chris van Swaay \textit{(CC BY-NC 4.0)} \minitab \scriptsize{\url{https://www.inaturalist.org/photos/18941543}} \\
Fig. 5: Top right, right & Jonathan Campbell \textit{(CC BY-NC 4.0)} \minitab \scriptsize{\url{https://www.inaturalist.org/photos/20120523}} \\
Fig. 5: Middle left, left & John Ratzlaff \textit{(CC BY-NC-ND 4.0)} \minitab \scriptsize{\url{https://www.inaturalist.org/photos/647514}} \\
Fig. 5: Middle left, right & Jessica \textit{(CC BY-NC 4.0)} \minitab \scriptsize{\url{https://www.inaturalist.org/photos/5595152}} \\
Fig. 5: Middle middle, left & i\_c\_riddell \textit{(CC BY-NC 4.0)} \minitab \scriptsize{\url{https://www.inaturalist.org/photos/1331149}} \\
Fig. 5: Middle middle, right & Pronoy Baidya \textit{(CC BY-NC-ND 4.0)} \minitab \scriptsize{\url{https://www.inaturalist.org/photos/5027691}} \\
Fig. 5: Middle right, left & Nicolas Olejnik \textit{(CC BY-NC 4.0)} \minitab \scriptsize{\url{https://www.inaturalist.org/photos/2006632}} \\
Fig. 5: Middle right, right & Carmelo López Abad \textit{(CC BY-NC 4.0)} \minitab \scriptsize{\url{https://www.inaturalist.org/photos/892048}} \\
Fig. 5: Bottom left, left & Luis Querido \textit{(CC BY-NC 4.0)} \minitab \scriptsize{\url{https://www.inaturalist.org/photos/13052253}} \\
Fig. 5: Bottom left, right & copper \textit{(CC BY-NC 4.0)} \minitab \scriptsize{\url{https://www.inaturalist.org/photos/22043211}} \\
Fig. 5: Bottom middle, left & vireolanius \textit{(CC BY-NC 4.0)} \minitab \scriptsize{\url{https://www.inaturalist.org/photos/13550702}} \\
Fig. 5: Bottom middle, right & Mathias D'haen \textit{(CC BY-NC 4.0)} \minitab \scriptsize{\url{https://www.inaturalist.org/photos/14943695}} \\
Fig. 5: Bottom right, left & tas47 \textit{(CC BY-NC 4.0)} \minitab \scriptsize{\url{https://www.inaturalist.org/photos/10691998}} \\
Fig. 5: Bottom right, right & Nik Borrow \textit{(CC BY-NC 4.0)} \minitab \scriptsize{\url{https://www.inaturalist.org/photos/13776993}} \\ \bottomrule
\end{tabular}
}
\end{table}

\end{document}